\documentclass[runningheads]{llncs}

\usepackage{graphicx}
\usepackage{comment}
\usepackage{amsmath,amssymb} 
\usepackage{color}
\usepackage{multirow}
\usepackage{pdfpages}

\begin{document}
\pagestyle{headings}
\mainmatter
\def\ECCVSubNumber{5105}  

\title{Detail Preserved Point Cloud Completion via Separated Feature Aggregation} 

\titlerunning{Detail Preserved Point Completion via Separated Feature Aggregation}
%
\author{Wenxiao Zhang\inst{1} \and
Qingan Yan\inst{2} \and
Chunxia Xiao\inst{1}\thanks{Corresponding author.}}
\authorrunning{Zhang et al.}
%
\institute{School of Computer Science, Wuhan University \\
\email{wenxxiao.zhang@gmail.com,cxxiao@whu.edu.cn} \and
JD.com American Technologies Corporation, CA\\
\email{qingan.yan@jd.com}}
\maketitle

\begin{abstract} 
Point cloud shape completion is a challenging problem in 3D vision and robotics. Existing learning-based frameworks leverage encoder-decoder architectures to recover the complete shape from a highly encoded global feature vector. Though the global feature can approximately represent the overall shape of 3D objects, it would lead to the loss of shape details during the completion process. In this work, instead of using a global feature to recover the whole complete surface, we explore the functionality of multi-level features and aggregate different features to represent the known part and the missing part separately. We propose two different feature aggregation strategies, named global \& local feature aggregation(GLFA) and residual feature aggregation(RFA), to express the two kinds of features and reconstruct coordinates from their combination. In addition, we also design a refinement component to prevent the generated point cloud from non-uniform distribution and outliers. Extensive experiments have been conducted on the ShapeNet dataset. Qualitative and quantitative evaluations demonstrate that our proposed network outperforms current state-of-the art methods especially on detail preservation. Our code is available at \url{https://github.com/XLechter/Detail-Preserved-Point-Cloud-Completion-via-SFA}.
\end{abstract}

\section{Introduction}
As the low-cost sensors like depth camera and LIDAR are becoming increasingly available, 3D data has gained large attention in vision and robotics community. However, viewpoint occlusion and low sensor resolution in 3D scans always lead to incomplete shapes, which can not be directly used in practical applications. To this end, it is desired to recover a complete 3D model from a partial shape, which has significant values in variety of tasks such as 3D reconstruction~\cite{dai2017shape}, robotics~\cite{varley2017shape}, scene understanding~\cite{dai2018scancomplete} and autonomous driving~\cite{qi2019deep}. 

Recent learning-based works succeed in performing shape completion on volumetric representation of 3D objects, such as occupied grids or TSDF volume, where convolution operations can be applied directly ~\cite{wu20153d,li2016shape,dai2017shape,han2017high,yang20173d}. However, volumetric representation always leads to expensive memory cost and low shape fidelity. In contrast, point cloud is a more compact and finer representation of 3D data than voxel but is harder to incorporate in the neural network due to its irregular properties. 
\begin{figure}[t]
\centering
\includegraphics[width=0.7\textwidth]{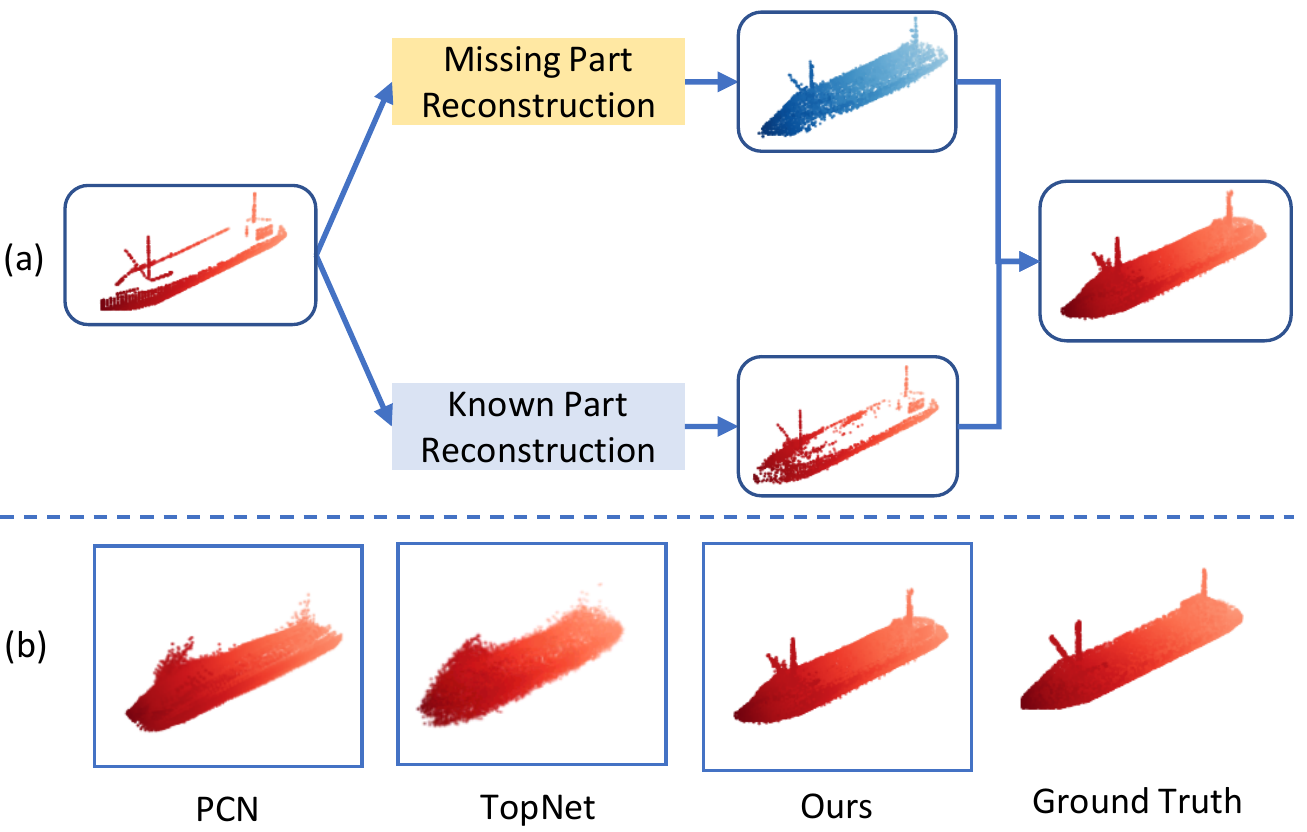}
\caption{(a ) Given a partial input, the proposed network reconstructs the complete shape from known and missing parts using separated feature representation. (b) Compared with PCN~\cite{yuan2018pcn} and TopNet~\cite{tchapmi2019topnet}, our method performs better in both detail preservation and latent shape prediction.}
\label{fig:intro}
\vspace{-0.6cm}
\end{figure}
Recently, several network architectures have been designed for direct point cloud shape completion, such as PCN~\cite{yuan2018pcn}, TopNet~\cite{tchapmi2019topnet}, RL-GAN-Net~\cite{sarmad2019rl}, demonstrating the advantage of point cloud shape completion through learning-based methods. These methods are all based on encoder-decoder networks, where the completed 3D model is recovered via a global feature vector obtained from the encoder. Though the encoded vector is able to represent the overall shape, merely considering it in shape completion will lead to the loss of geometric details existing in original point clouds.

To address this problem, rather than using only a global feature vector to generate the whole complete model, we extract multi-level features and aggregate different-level features to represent the known and missing parts separately, which contributes to both existing detail preservation and missing shape prediction, as illustrated in Figure~\ref{fig:intro}(a).

We first extract multi-level features for each point by a hierarchical feature learning architecture. These features can present the existing details well, but they do not involve enough missing part cues. If we directly use these features for shape completion, the generated results are likely tangled into the existing partial input. To this end, we consider aggregating diverse features for the known and missing parts separately. We propose two feature aggregation strategies, \emph{i.e., global $\&$ local feature aggregation(GLFA)} and \emph{residual feature aggregation(RFA)}. For \emph{global $\&$ local feature aggregation}, the main idea is that we leverage local features to represent the known part while aggregate global features for missing part. For \emph{residual feature aggregation}, we aim to represent the missing part with residual features which indicates the difference between the holistic shape and the partial model. We then combine the known and missing part features with a proper ratio and reconstruct the complete model from the combined features. Finally, a successive refinement component is designed to converge the generated model to be uniformly distributed and reduce the outliers.

To summarize, our main contributions are:
\begin{itemize}
    \item We propose a novel learning based point cloud completion architecture which achieves better performance on both detail preservation and latent shape prediction by separated feature aggregation strategy.
    \item We design a refinement component which can refine the reconstructed coordinates to be uniformly distributed and reduce the noises and outliers.
    \item  Extensive experiments demonstrate our proposed network outperforms state-of-the-art 3D point cloud completion methods both quantitatively and qualitatively.
\vspace{-0.15cm}
\end{itemize}

\section{Related Work}
\vspace{-0.15cm}
\noindent{\bfseries Deep learning on point cloud.}  Qi {\em et al.}~\cite{qi2017pointnet} first introduced a deep learning network PointNet which uses symmetric function to directly process point cloud. PointNet++~\cite{qi2017pointnet++} captures point cloud local structure from neighborhoods at multiple scales. FoldingNet~\cite{yang2018foldingnet} uses a novel folding-based decoder which deforms a canonical 2D grid onto the underlying shape surface. A series of network architectures on point cloud have been proposed in succession for point cloud analysis ~\cite{su2018splatnet,hua2018pointwise,li2018s,wang2018deep,le2018pointgrid,xu2018spidercnn,wang2018local,rethage2018fully,liu2019relation,wu2019pointconv,he2019geonet,zhao20193d,zhao2019pointweb,chen2019fast}, and related applications such as object detection~\cite{dinesh2018carfusion,qi2018frustum,wang2019pseudo,shi2019pointrcnn,lang2019pointpillars,giancola2019leveraging,qi2019deep} and reconstruction \cite{wang2018local,yang2019pointflow,han2019multi}. 

\noindent{\bfseries Non-learning based shape completion.} Shape completion has long been a popular problem on interest in the graphics and vision field. Some effective descriptors have been developed in the early years, such as~\cite{nealen2006laplacian,sorkine2004least,kazhdan2013screened}, which leverages geometric cues to fill the missing parts in the surface. These methods are usually limited to fill only the small holes.  Another way to complete the shape is to find the symmetric structure as priors to achieve the completion~\cite{thrun2005shape,pauly2008discovering,mitra2006partial}. However, these methods work well only when the missing part can be inferred from the existing partial model. Some researchers proposed data-driven methods~\cite{li2015database,shi2016data,kim2012acquiring} which usually retrieve the most likely model based on the partial input from a large 3D shape database. Though convincing results can be obtained, these methods are time-consuming in matching process according to the database size. 

\noindent{\bfseries Learning based shape completion.} Recently more researchers tend to solve 3D vision tasks using learning methods. Learning based methods on shape completion usually use deep neural network with a encoder-decoder architecture to directly map the partial input to a complete shape. Most pioneering works \cite{wu20153d,li2016shape,dai2017shape,han2017high,yang20173d} rely on volumetric representations where convolution operations can be directly applied. Volumetric representations lead to large computation and memory costs, thus most works operate on low dimension voxel grids leading to details missing. To avoid these limitations, Yuan {\em et al.} proposed PCN~\cite{yuan2018pcn} which directly generates complete shape with partial point cloud as input. PCN recovers the complete point cloud in a two-stage process which first generates a coarse result with low resolution and then recovers the final output using the coarse one. TopNet~\cite{tchapmi2019topnet} explores hierarchical rooted tree structure as decoder to generate arbitrary grouping of points in completion task. RL-GAN-Net~\cite{sarmad2019rl} presents a completion framework using reinforcement learning agent to control the GAN generator. All these approaches generate the complete point cloud by feeding a encoded global feature vector to a decoder network. Though the global feature can almost represent the underlying holistic surface, it discards many details existing in the partial input.


\begin{figure*}[t]
\centering
\includegraphics[width=0.95\textwidth]{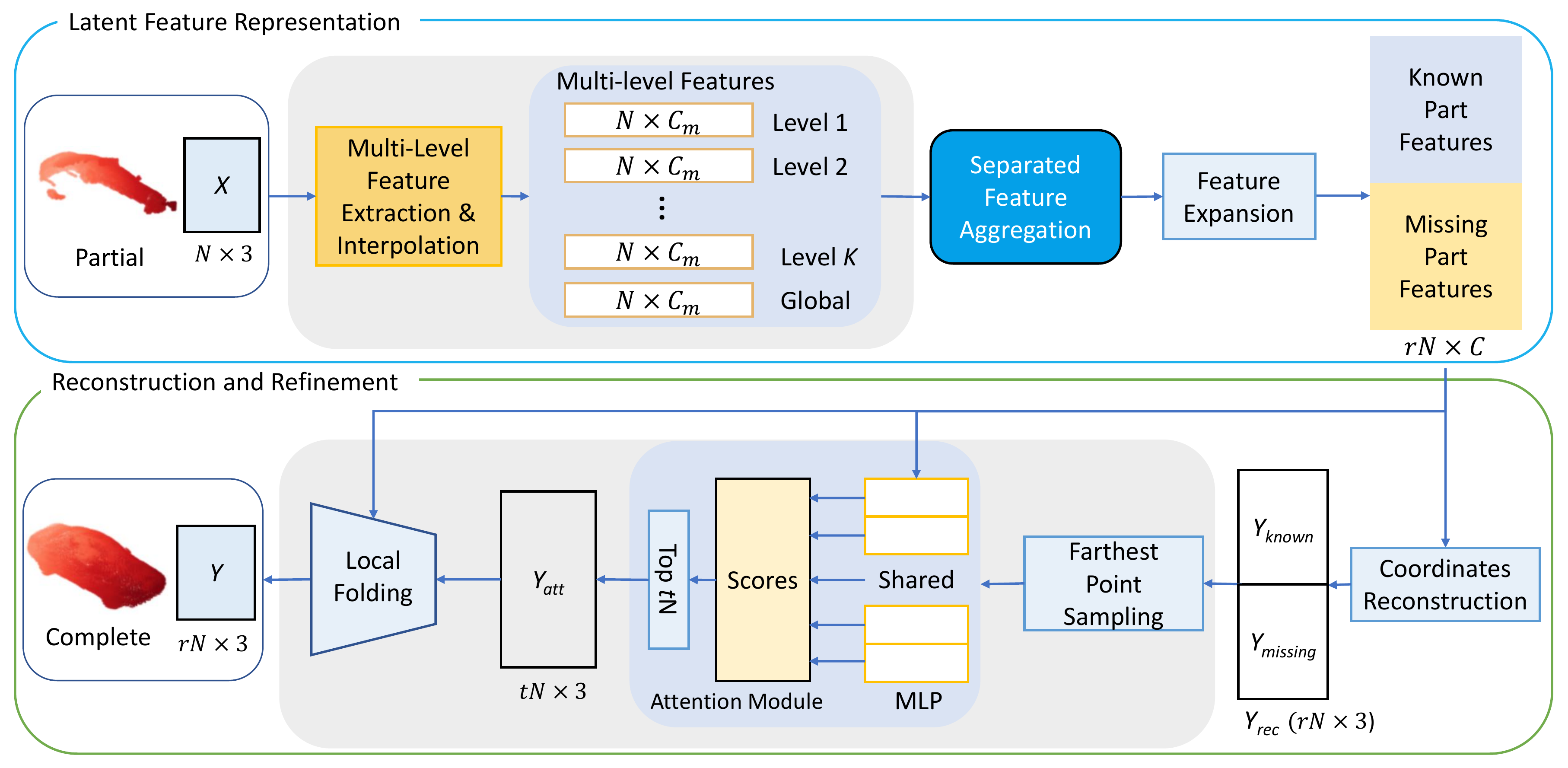}
\caption{Overall network architecture. Given a partial point cloud of N points with
XYZ coordinates, our network extracts multi-level features and interpolates each level feature to the same size $N \times C_m$ (implemented with PointNet++ ~\cite{qi2017pointnet++} layers). Multi-level features are then aggregated to represent known and missing parts separately with feature aggregation strategy. A high-resolution completion result $Y_{rec}$ of size $rN \times 3$ is reconstructed from the expanded features where $r$ is the expansion factor. The reconstructed results are finally refined by a refinement component.}
\label{fig:overview}
\vspace{-0.5cm}
\end{figure*}

\vspace{-0.3cm}
\section{Network Architecture}
\vspace{-0.25cm}

The overall architecture of our network is shown in Figure~\ref{fig:overview}. First, the multi-level features are extracted via a hierarchical feature learning to efficiently represent both local and global properties. Then we aggregate features for known and missing parts of the point set separately to provide explicit cues for detail preservation and shape prediction. After that, we expand the features and reconstruct the coordinates from the expanded features to obtain a high-resolution result. Finally, we design a refinement component to make the complete point cloud uniformly distributed and smooth.

\vspace{-0.15cm}
\subsection{Multi-level Features Extraction}

Both local and global structures can be efficiently explored by extracting features from different levels. We extract multi-level features via a hierarchical feature learning architecture proposed in PointNet++~\cite{qi2017pointnet++}, which is also used in PU-Net\cite{yu2018pu}. The feature extraction module consists of multiple layers of sampling and grouping operation to embed multi-level features and then interpolates each level feature to have the same point number and feature size of $N \times C_m$, where $N$ is the input point number and $C_m$ refers to the interpolated multi-level feature dimension. We progressively capture features with gradually increasing grouping radius. The features at the $i$-th level is defined as $f_{i}$. Different from the PU-Net~\cite{yu2018pu}, we extract a global feature $f_{global}$ from the last-level feature. The global feature $f_{global}$ have the same size $1 \times C_m$ and we duplicate it to $N \times C_m$. 



\vspace{-0.25cm}
\subsection{Separated Feature Aggregation}
\label{sec:RFA}
If we directly concatenate the multi-level features for completion, a main problem is that these features can preserve the existing details, but it does not capture enough information to complete the missing shape. The generated points are easier to be tangled at the original partial input even if we add a repulsion loss on the generated results. We discuss this problem in experiments (Section ~\ref{sec:wofa}).

To this end, we consider providing explicit cues for the network to balance detail preservation and shape prediction. We separately represent the known and missing parts with diverse features. For known part features, denoted as $f_{known}$, we want to ensure the propagation of local structure of the input. The missing part features, denoted as $f_{missing}$, can be regarded as the features predicted from global structure, or features indicating the difference between the complete shape and the existing partial structure. 

\begin{figure}[t]
\centering
\includegraphics[width=0.7\textwidth]{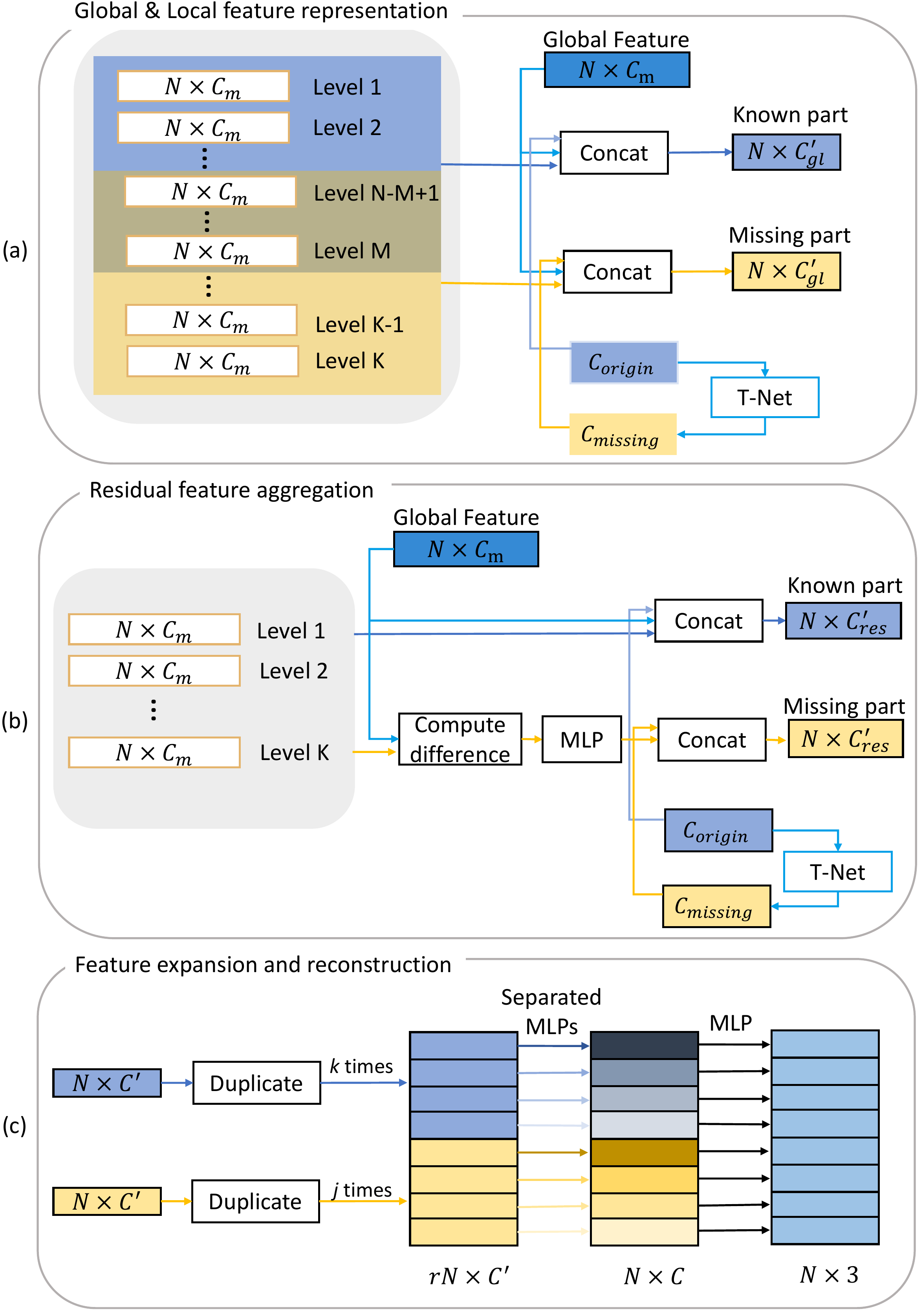}
\caption{Illustration of the proposed feature aggregation strategies and the process of feature expansion and reconstruction.}
\label{fig:RFA}
\vspace{-0.4cm}
\end{figure}

Taking the advantages of multi-level features, we propose and compare two different types of feature aggregation strategies namely GLFA and RFA, which is illustrated in Figure~\ref{fig:RFA} (a)(b).

\noindent{\bfseries Global \& Local Feature Aggregation (GLFA).}
Intuitively, the known part should be represented with more local features to keep the original details, whereas the missing part should be expressed with a more global feature to predict the latent underlying surface for the completion task. Generally, lower-level features layers in a network correspond to local features in smaller scales, while higher level features are more related to the global features. To this end,  we represent the $f_{known}$ and $f_{missing}$ as:
\begin{equation} 
\begin{split}
f_{known}&=[f_1, \ldots, f_{m}, C_{origin}, f_{global}],\\
f_{missing}&= [f_{n-m+1}, \ldots, f_n, C_{missing}, f_{global}],\\
C_{missing}& = \mathcal{T}(C_{origin}),\\
m&= {\lfloor {\frac{n}{2}} \rfloor} + 1,\\
\end{split}
\end{equation} 
where $n$ is the total level number. We aggregate the first $m$ levels features $[f_1, \ldots, f_m]$ to $f_{known}$ and last $m$ level features $[f_{n-m+1}, \ldots, f_n]$ to $f_{missing}$.

$C_{origin}$ is the original points coordinates in the partial point cloud. $C_{missing}$ indicates the possible coordinates of the missing part. At first, we just use the $C_{origin}$ as $C_{missing}$, but we find the network can converge faster if we set $C_{missing}$ with more proper initial coordinates. To this end, we leverage a learnable transformation net $\mathcal{T}(\cdot)$ to transform $C_{origin}$ to proper coordinates as $C_{missing}$. $\mathcal{T}(\cdot)$ is similar to the T-Net proposed in Pointnet\cite{qi2017pointnet}, containing a series of multi-layer perceptrons, a predicted transform matrix and coordinates bias. Our observation for using a transformation net is the symmetry properties of most of objects in the world, thus transforming the original coordinates properly can better fit the missing part coordinates. 



\noindent{\bfseries Residual Feature Aggregation (RFA).}
A more direct way to consider this issue is to represent the missing part with features which can be seen as residual features between the global shape and the known part. We first compute the difference between the global feature vector and known part features in feature space. After that, a successive shared multi-layer perceptron is applied to generalize the difference to the residual features in the latent feature space.
The $f_{known}$ and $f_{missing}$ are represented as:
\begin{equation} 
\begin{split}
f_{known}&= [f_1, \ldots, f_n, C_{origin}, f_{global}],\\
f_{missing}&= [f_{res_i} , \ldots, f_{res_n}, C_{missing}, f_{global}],\\
C_{missing}& = \mathcal{T}(C_{origin}),\\
f_{res_i}&=\mathbf{MLP}([f_{global} - f_i]),
\end{split}
\end{equation} 
where $f_{res_i}$ indicates the residual features between $f_{global}$ and the $f_i$, and $\mathbf{MLP}(\cdot)$ refers to a shared MLP of several fully connected layers. 

\subsection{Feature Expansion and Reconstruction}

\noindent{\bfseries Feature expansion with combination.}
After feature aggregation, we combine both the known and missing features and expand the number of features to high-resolution representation. As shown in Figure~\ref{fig:RFA}(c), we expand the feature number by duplicating $f_{known}$ and $f_{missing}$ for $j$ and $k$ times respectively, where $j+k=r$. We then apply $r$ separated MLPs to the $r$ feature sets to generate diverse point sets. Every MLP shares weights in the feature set.

An important factor is how we define the combination ratio of $j$ and $k$. In our experiments, we define $j:k=1:1$, as the visible part is often about a half of the whole object. We discuss the effect of the ratio in experiments part (Section ~\ref{sec:experiments}).

\noindent{\bfseries Coordinates reconstruction.} We reconstruct a coarse model $Y_{rec}$ from the expanded features via a sharded MLP and the output is point coordinates $rN\times 3$. The corresponding reconstructed coordinates from the known part and missing
part are denoted as $Y_{known}$ and $Y_{missing}$.

\vspace{-0.25cm}
\subsection{Refinement Component}
In practice, we notice that the reconstructed points are easy to locate too close to each other due to the high correlation of the duplicated features, and suffer from noises and outliers. Thus, we design a refinement component to enhance the distribution uniformity and reduce useless points.

We first apply farthest point sampling (FPS) to get a relatively uniformly subsampled $\frac{r}{2}N$ point set. However, FPS algorithm is random and the subsampling depends on which point is selected first, which will not get rid of the noises and outliers in the point set. 

We apply a following attention module to reduce the incorrect points. The attention module comprises a shared MLP of 3 stacking fully connected layers with softplus activation function applied on the last layer to produce a score map. It selects the top $tN$ features and points as shown in Figure~\ref{fig:overview}. The input to the attention module is the corresponding $\frac{r}{2}N \times C$ features to the selected points, and the output is the indexes of the most significant $tN$ points. The selected points are denoted as $Y_{att}$.

Finally, we apply a local folding unit~\cite{yang2018foldingnet} to approximate a smooth surface of high resolution, which is proposed in PCN~\cite{yuan2018pcn}. The input to the local folding unit is the $tN \times (3+C)$ points and corresponding features. For each point, a patch of $u^2$ points is generated in the local coordinates centered at ${x_i}$ via the folding operation, and transformed into the global coordinates by adding ${x_i}$ to the output. The final output is the refined point coordinates $rN\times 3$.


\subsection{Loss Function}
To measure the differences between two point clouds $(S_{1},S_{2})$, Chamfer Distance (CD) and Earth Mover’s Distance (EMD) are introduced  in recent work ~\cite{fan2017point}.
We choose the Chamfer distance like \cite{tchapmi2019topnet,sarmad2019rl}, due to its efficiency over EMD.
\begin{small}
\begin{equation} 
\mathcal{L}_{CD}(S_{1},\!S_{2})\!=\!\frac{1}{\vert S_{1}\vert }\sum_{x\in S_{1}}\!\min_{y\in S_{2}}\Vert x-y\Vert_{2}
+\frac{1}{\vert S_{2}\vert }\sum_{y\in S_{2}}\!\min_{x\in S_{1}}\Vert y-x\Vert_{2}. 
\end{equation}
\end{small}

Besides FPS algorithm, we also explicitly ensure the distribution uniformity of the attention module output in the loss function to reduce the incorrect points. We apply the repulsion loss proposed in PU-Net~\cite{yu2018pu} which is defined as:

\begin{equation}
\mathcal{L}_{rep}(S)=\sum_{x_i\in S}\sum_{x_{i^{\prime}}\in K(x_i)}\eta(\Vert x_{i^{\prime}}-x_{i}\Vert)w(\Vert x_{i^{\prime}}-x_{i}\Vert),
\end{equation}
where $\eta(\cdot)$ and $w(\cdot)$ are two repulsion term to penalize $x_{i}$ if $x_{i}$ is too close to its neighboring point $x_{i^{\prime}}$.

We jointly train the network by minimizing the following loss function:
\begin{equation}
\begin{split}
{L}_{sum}=\alpha \mathcal{L}_{CD}(Y_{rec}, {Y}_{gt}) + \mathcal{L}_{CD}(Y_{att},{Y}_{gt})\\
 + \mathcal{L}_{CD}(Y_{final}, {Y}_{gt}) + \beta \mathcal{L}_{rep}(Y_{att}),
\end{split}
\end{equation}
where we set $\alpha =0.5$ as we do not need $Y_{rec}$ to be an very accurate result, and $\beta = 0.2$ in our experiments. Note that we apply repulsion loss only on $Y_{att}$, as local folding unit can approximate a smooth surface
which represents the local geometry of a shape, so we just need to ensure $Y_{att}$ to be uniformly distributed.

We do not explicitly constrain $Y_{known}$ and $Y_{missing}$ in the loss function to be close to the known part and missing points respectively.  If we supervise $Y_{known}$ with partial input, it will leave the whole completion task to the missing part features reconstruction. We expect the network itself to learn to reconstruct diverse coordinates according to the different feature representation for known and missing parts, which makes the completion can process in both $Y_{known}$ and $Y_{missing}$. Our experiments demonstrate this in Sec \ref{sec:visualres}.


\section{Experiments}

\begin{table*}[t]
 \centering
 \begin{tabular}{c|c|c|c|c|c|c|c|c|c}
    \hline
    Method & airplane& cabinet& car& chair&lamp& sofa& table& vessel& Average\\
    \hline
FC&5.69&11.02&8.77&10.96&11.13&11.75&9.32&9.72&9.79\\
Folding&5.96&10.83&9.27&11.24&12.17&11.63&9.45&10.02&10.07\\
PCN&5.50&10.625&8.69&10.99&11.33&11.67&8.59&9.66&9.63\\
TopNet  &5.85 &10.78 &8.84 &10.80 &11.15 &11.41 &8.79 &9.17 &9.60\\
    \hline
   NSFA-RFA  &\textbf{4.76} &\textbf{10.18} &\textbf{8.63} &\textbf{8.53} &\textbf{7.03} &10.53 &7.35 &7.48 &\textbf{8.06}\\
    NSFA-GLFA &4.85 &10.31 &8.92 &8.99 &7.24 &\textbf{10.28} &\textbf{7.33} &\textbf{7.15} &8.14\\ 
    \hline
  \end{tabular}
  \caption{Quantitative comparison on known categories with state-of-the-art methods with the metric as Chamfer Distance multiplied by $10^4$.}
  \label{table:comparision_known}
\vspace{-0.3cm}
\end{table*}

\begin{table*}[t]
 \centering
 \begin{tabular}{c|c|c|c|c|c|c|c|c|c|c}
    \hline
\multirow{2}*{Method}&\multicolumn{5}{|c|}{Similar}&\multicolumn{5}{|c}{Dissimilar}\\
\cline{2-11}
     & bus&bed &bookshelf &bench& Avg&guitar &motor&skateboard&pistol&Avg \\
    \hline
    FC  &9.82 &21.23 &15.12 &10.81&14.20 &9.92 &14.56 &12.00 &14.97 &12.90\\
    Folding  &10.58 &19.08 &14.88 &10.55&13.80  &9.06 &15.56 &11.91 &13.13 &12.40\\ 
    PCN  &9.46 &21.63 &14.79 &11.02&14.20 &10.40 &14.75 &12.04 &14.23 &12.90 \\
    TopNet  &9.31 &20.38 &14.12 &10.16 &13.40 & 9.88 &14.30 &9.26 &12.86 &11.50\\ 
    \hline
    NSFA-RFA &9.43 &18.21 &12.50 &9.83 &12.40 &
\textbf{7.49} &11.41 &9.09 &\textbf{9.09} &\textbf{9.20}\\
NSFA-GLFA &\textbf{9.26} &\textbf{15.43} &\textbf{11.92} &\textbf{9.26} &\textbf{11.40}&7.71 &\textbf{9.94} &\textbf{9.06}&10.16&\textbf{9.20}\\
    \hline
  \end{tabular}
  \caption{Quantitative comparison on novel categories with state-of-the-art methods with the metric as Chamfer Distance multiplied by $10^4$.
  }
  \label{table:comparision_novel}
\vspace{-0.7cm}
\end{table*}
\noindent{\bfseries Training set.} We conduct our experiments on the dataset used by PCN~\cite{yuan2018pcn} which is a subset of the Shapenet dataset. The ground truth  contains 16384 points uniformly sampled from the mesh, and the partial inputs with 2048 points are generated by back-projecting 2.5D depth images into 3D.
The training set contains 28974 different models from 8 categories. Each model contains a complete point cloud with about 7 partial point clouds taken from different viewpoint for data augmentation. The validation set contains 100 models. 

\noindent{\bfseries Testing set from ShapeNet.} 
The testing set from ShapeNet\cite{chang2015shapenet} are divided into two sets: one contains 8 known object categories on which the models are trained; another contains 8 novel categories that are not in the training set. The novel categories are also divided into two groups: one that is visually similar to the known categories, and another that is visually dissimilar. There are 150 models in each category.

\noindent{\bfseries Testing set from Kitti.} 
We also test our methods on real-world scans from Kitti dataset \cite{geiger2013vision}. The testing scans are cars which are extracted from each frame according to the ground truth object bounding boxes. The testing set contains 2483 partial point clouds labeled as cars. 


\noindent{\bfseries Training setup.} 
We train our model for 30 epochs with a batch size of 8. The initial learning rate is set to be 0.0007 which is decayed by 0.7 for every 50,000 iterations. We set other parameters $r=8$, $t=2$, $u=2$ in our implements.

\begin{figure*}[t]
\centering
\includegraphics[width=0.975\textwidth]{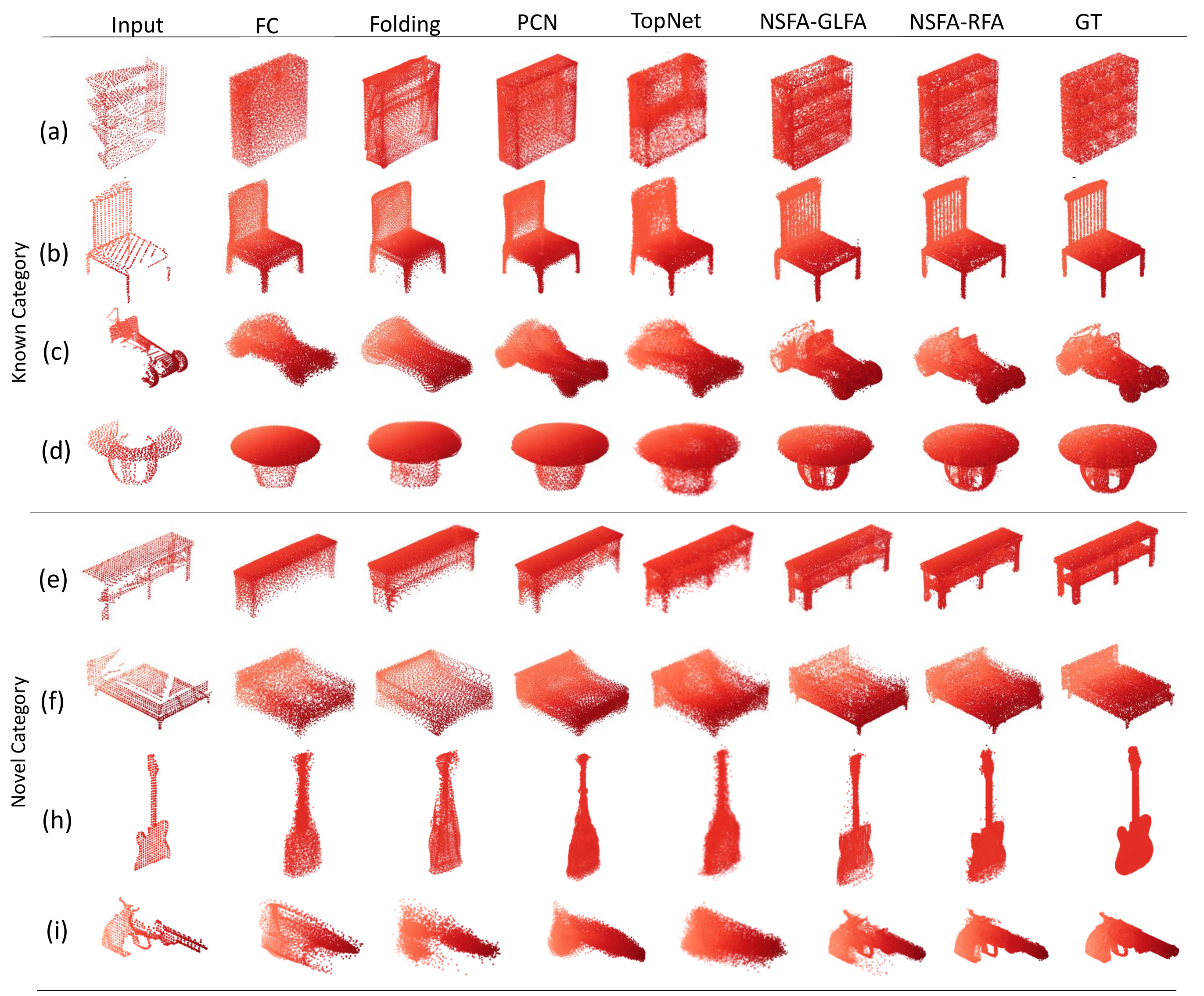}
\caption{Qualitative comparisons on both known and novel categories.}
\vspace{-0.3cm}
\label{comparison}
\end{figure*}

\vspace{-0.5cm}
\subsection{Completion Results on ShapeNet}
We qualitatively and quantitatively compare our network on both known categories and novel categories with several state-of-the-art point cloud completion methods: FC (Pointnet auto-encoder)~\cite{qi2017pointnet}, FoldingNet~\cite{yang2018foldingnet}, PCN~\cite{yuan2018pcn}, and TopNet~\cite{tchapmi2019topnet}. For  FC, Folding, PCN, we use the pre-trained model and evaluation codes released in the public project of PCN on github. For TopNet, we use their public code and
retrain their network using the training set in PCN. For the sake of clarity, we denote our network with separated feature aggregation of GLFA and RFA strategy respectively as NSFA-GLFA and NSFA-RFA. We choose our network of 5 levels during feature extraction as the baseline network. Table~\ref{table:comparision_known} and ~\ref{table:comparision_novel} show the quantitative comparison results on known and novel categories respectively. Both NSFA-GLFA and NSFA-RFA achieve lower values for the evaluation metrics in both known and novel categories. 

Besides quantitative results, as we claim that our methods can preserve the shape details, we select some qualitative examples of models with much details from the testing set. Results are shown in Figure~\ref{comparison} where (a)-(d) are from known categories and (e)-(i) are from novel categories. We can observe that FC, Folding, PCN and TopNet can predict a overall shape but most details of the model are lost. In contrast, NSFA-GLFA and NSFA-RFA show their outperformance in both detail preservation and missing shape prediction.

\vspace{-0.5cm}
\subsection{Completion Results on Kitti}
\vspace{-0.3cm}
As there are no complete ground truth point clouds for Kitti, we use two metrics proposed in PCN to quantitatively evaluate the performance: 1) Fidelity error, which is the average distance from each point in the input to its nearest neighbour in the output. This measures how well the input is preserved; 2) Minimal Matching Distance (MMD), which is the Chamfer Distance between the output and the car point cloud from ShapeNet that is closest to the output point cloud in terms of CD. This measures how much the output resembles a typical car. 

\begin{figure}[h]
  \vspace{-1.2cm}
\centering
\includegraphics[width=0.8\textwidth]{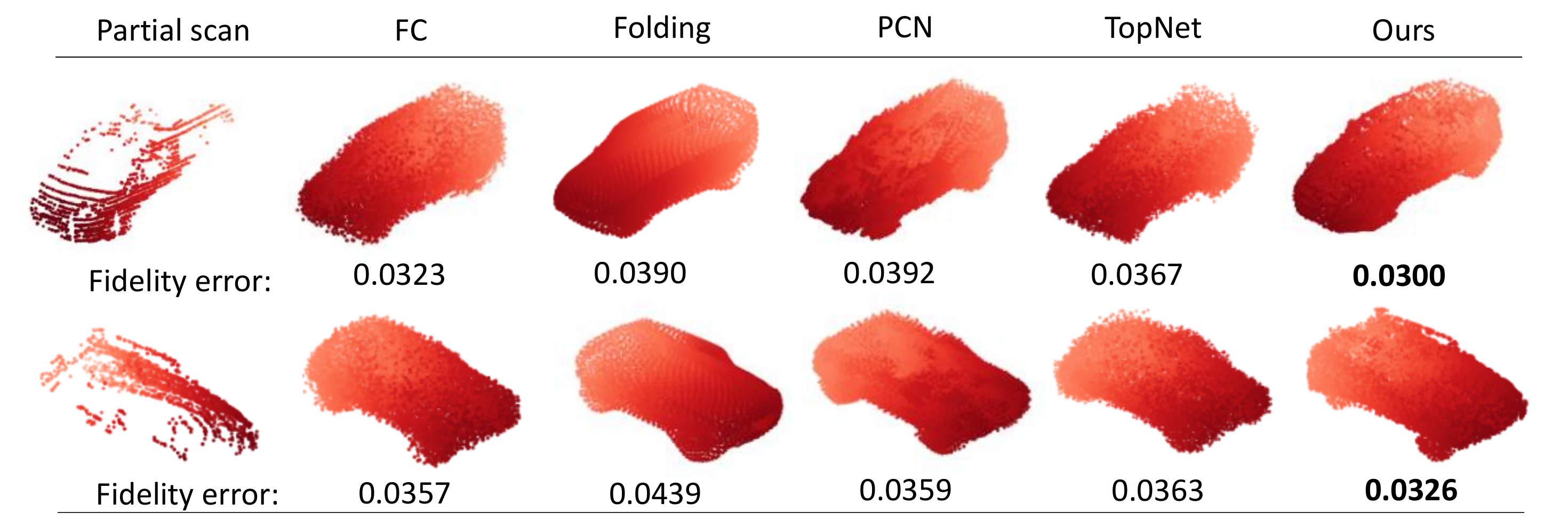}
  \vspace{-0.3cm}
\caption{Results on Kitti dataset. The numbers below the completion results show the fidelity error for each model.}
\label{fig:kitti}
\end{figure}

The quantitative and qualitative results are shown in Table \ref{table:kitti} and Figure \ref{fig:kitti} respectively. The fidelity error of each model is also attached in qualitative results. We use NSFA-RFA as our method for comparison as NSFA-RFA performs better than NSFA-GLFA on car category on ShapeNet. 

From both quantitative and qualitative results, it can be observed that  our method achieves lowest fidelity error, which meets our claim that our network can preserve the existing details better. In Figure \ref{fig:kitti}, we can see other methods also present reasonable results, but some details in the partial model seems to be lost. Besides, the performances of all the methods on MMD metric are very close, indicating the results from all methods can resemble typical cars.

\begin{table*}[h]
\vspace{-0.5cm}
 \centering
 \begin{tabular}{c|c|c|c|c|c}
    \hline
    Method & FC& Folding& TopNet& PCN&Ours\\
    \hline
    Fidelity error&0.0331&0.0361&0.0308&0.0335&\textbf{0.0261}\\
    \hline
    MMD &0.0148 &\textbf{0.0146} &0.0158&0.0151&0.0154\\
    \hline
  \end{tabular}
  \caption{Quantitative comparison on known categories with state-of-the-art methods with the metric as Chamfer Distance.}
  \label{table:kitti}
\vspace{-0.2cm}
\end{table*}

  \vspace{-1.0cm}
\subsection{Reconstructed Coordinates Visualization}
\label{sec:visualres}
\begin{figure*}[t]
\centering
\includegraphics[width=0.9\textwidth]{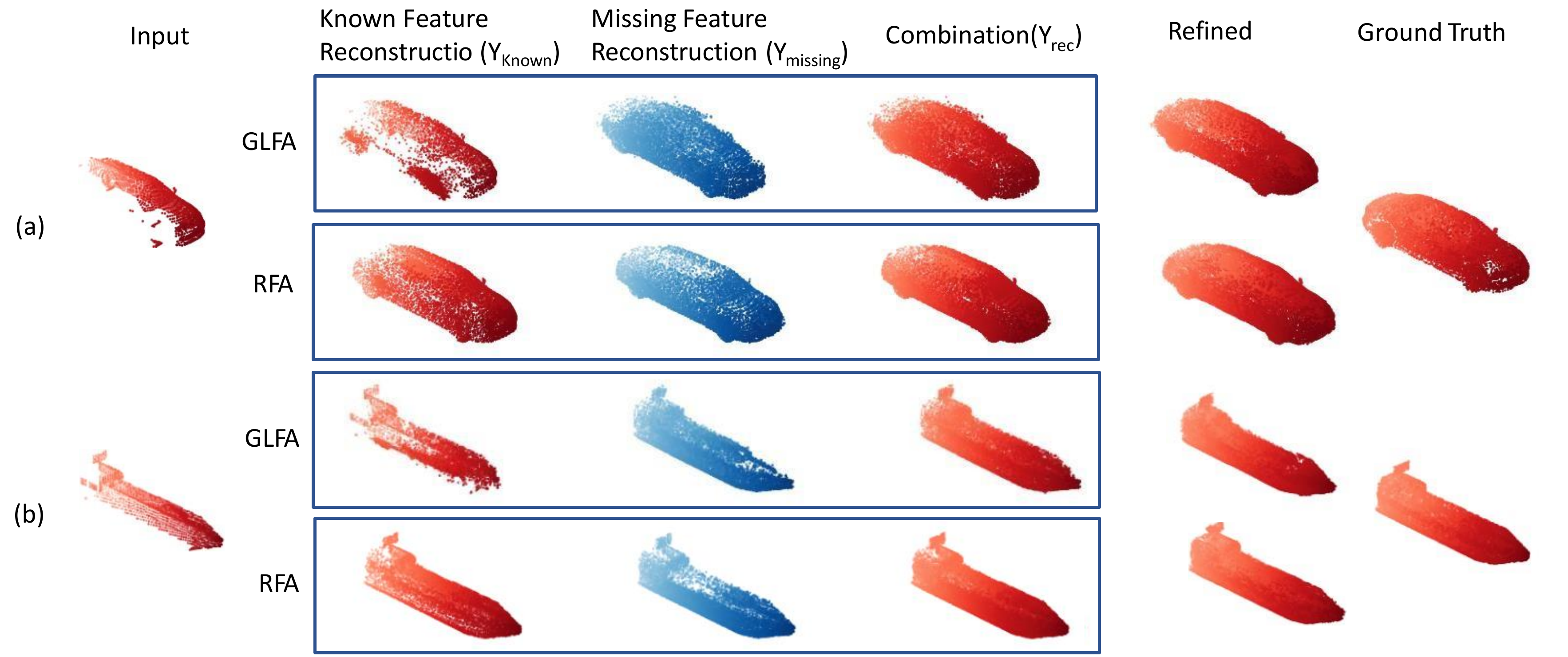}
\caption{Reconstructed coordinates visualization from known and missing part features.}
\label{fig:respective}
\vspace{-0.3cm}
\end{figure*}

As we consider the known and missing parts separately, we visualize the reconstructed coordinates $Y_{known}$ and $Y_{missing}$ in $Y_{rec}$ from the known part and missing part features to validate our method. Figure~\ref{fig:respective} shows the visualization examples of both NSFA-GLFA and NSFA-RFA. In general, $Y_{known}$ is closer to the partial input, and $Y_{missing}$ is closer to the missing shape. Meanwhile, there are also completion effects for the $Y_{known}$, and $Y_{missing}$ also has some original points. This is reasonable as we aggregate global feature to both part features, so the completion can progress in both parts. Specifically, the completion effects on $Y_{known}$ of NSFA-RFA is more significant than NSFA-GLFA. The reason may be that we concatenate all level features to $f_{known}$ of NSFA-RFA but only low-level local features in NSFA-GLFA, thus the completion effects are more apparent on the $Y_{known}$ of NSFA-RFA. 

\vspace{-0.3cm}
\subsection{Feature Aggregation Strategy Evaluation}
\begin{figure}[t]
\centering
\includegraphics[width=0.65\textwidth]{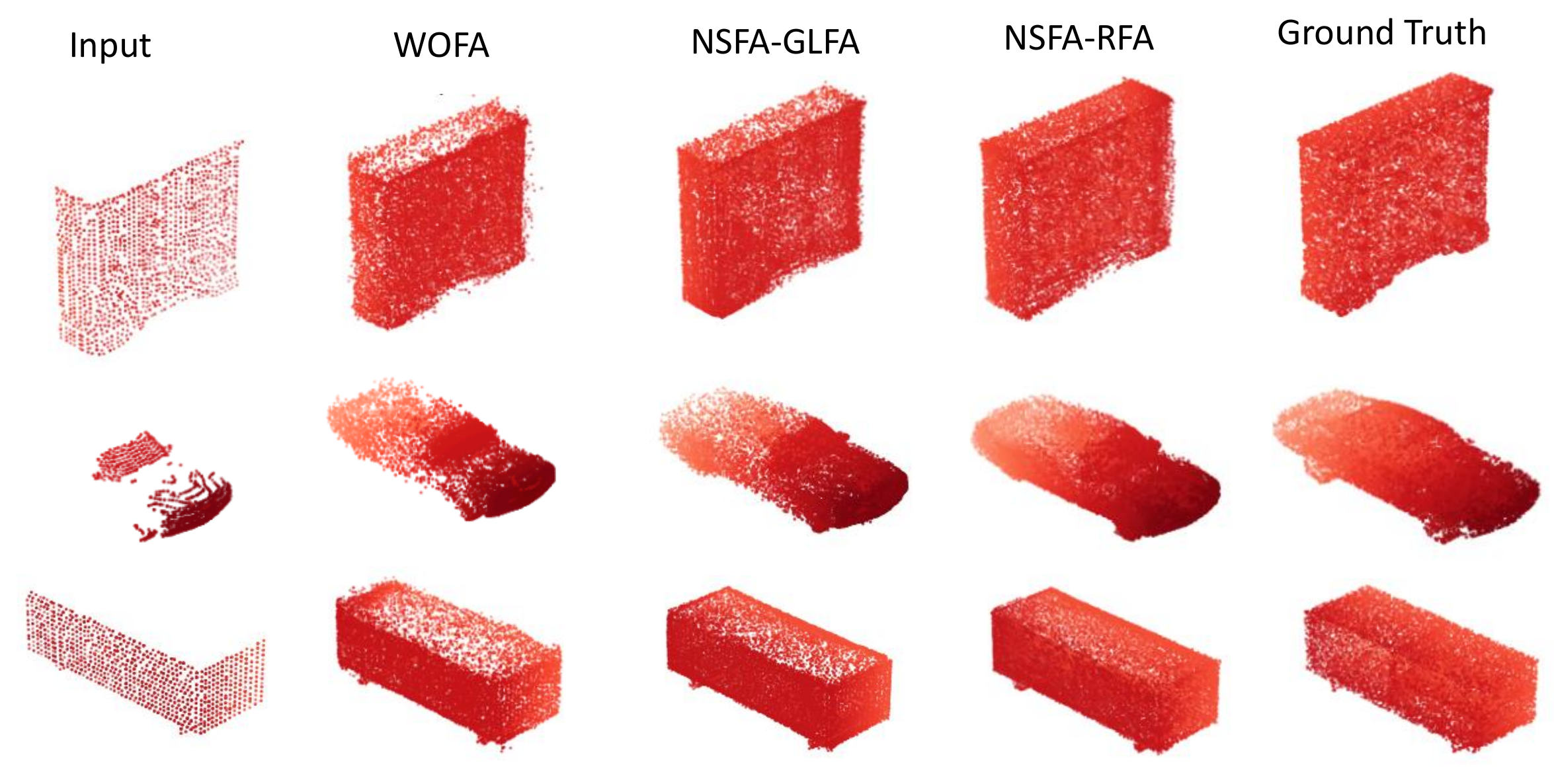}
\caption{Qualitative evaluation of the proposed feature aggregation strategies.}
\label{fig:wofa}
\vspace{-0.25cm}
\end{figure}

\vspace{-0.05cm}
\label{sec:wofa}
As we mentioned in Section ~\ref{sec:RFA}, directly concatenating the multi-level features for completion will lead to imbalance between detail preservation and shape completion. We denote our network without feature aggregation strategies as NWoSFA. We evaluate the proposed feature aggregation strategies with qualitative examples shown in Figure~\ref{fig:wofa}. It can be seen that the generated points of  NWoSFA are more likely to be gathered at the original input. This effect is released in NSFA-GLFA and the generated points of NSFA-RFA seems can be well spread on the surface. The reason why NSFA-RFA performs better than NSFA-GLFA may due to its more significant completion effects on the known part as mentioned in Section ~\ref{sec:visualres}. The completion effects on the known part can help fill the missing shape better. The quantitative results are shown in Table 5. It is notable that NSFA-RFA and NSFA-GLFA significantly boost the performance on the known categories. For novel categories, NSFA-GLFA achieves the lowest Chamfer Distance but NSFA-RFA seems not work well. This indicates that it is hard to generate missing part features using the residual feature aggregation for a totally unseen category. On contrast, NSFA-GLFA uses the global feature to represent the missing part, which is more suitable for novel category.


\begin{table}[h]
\vspace{-0.25cm}
\begin{minipage}[t]{.6\linewidth} 
\centering 
 \begin{tabular}{c|c|c|c|c}
\hline
\multirow{2}*{Method}&\multicolumn{2}{|c}{NSFA-RFA}&\multicolumn{2}{|c}{NSFA-GLFA} \\
\cline{2-5}
&Known&Novel&Known&Novel\\
    \hline
    Without Refine &8.65&10.09&9.38&10.70\\
    \hline
    With Refine &\textbf{8.06}&\textbf{10.08}&\textbf{8.14}&\textbf{9.98}\\
    \hline
  \end{tabular}
\label{table:worefine}
  \caption{Quantitative evaluation of the propos-\protect\\ed refinement component on known and novel \protect\\ categories.} 
\end{minipage}%
\begin{minipage}[t]{.4\linewidth} 
\centering 
 \begin{tabular}{c|c|c}
\hline
Method&Known&Novel \\
    \hline
    NWoSFA &8.74&10.02\\
    \hline
    NSFA-RFA &\textbf{8.06}&10.08\\
    \hline
    NSFA-GLFA &8.14&\textbf{9.98}\\
    \hline
  \end{tabular}
\label{table:wofa}
  \caption{Quantitative evaluation of the proposed feature aggregation strategies.} 
\end{minipage} 
\vspace{-0.35cm}
\end{table}

\vspace{-0.2cm}
\subsection{Effectiveness of Refinement Component}
\begin{figure*}[t]
\centering
\includegraphics[width=0.85\textwidth]{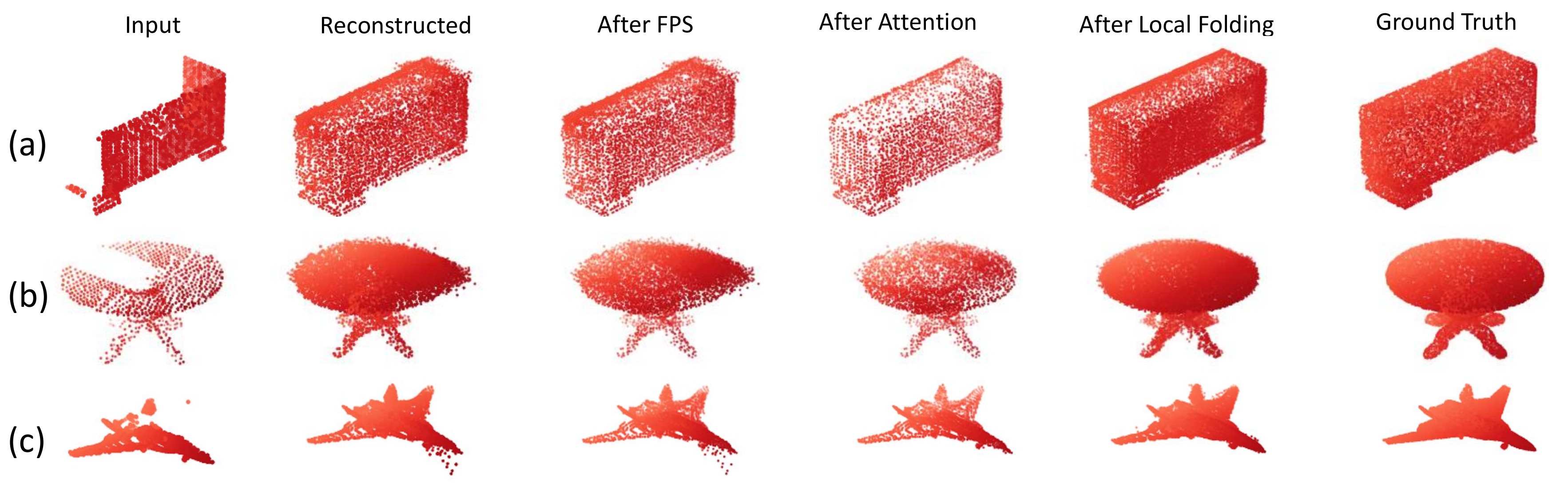}
\caption{Effectiveness of the refinement component. FPS module makes the result to be more uniformly distributed and the attention module reduces the invalid points. The local folding unit spread the points to a smooth surface.}
\vspace{-0.5cm}
\label{fig:refine}
\end{figure*}

We also evaluate the effectiveness of the refinement component. Some intermediate results produced by NSFA-GLFA are shown in Figure~\ref{fig:refine}. The FPS algorithm enhances the point set distribution uniformity, but it can not get rid of the incorrect points. With the attention module, the generated points are more uniformly distributed with fewer outliers and noises. Finally, the local folding unit spread the points to a high-resolution result with smooth surface. The quantitative results are also shown in Table 4.

\begin{figure}[t]
\centering
\includegraphics[width=0.8\textwidth]{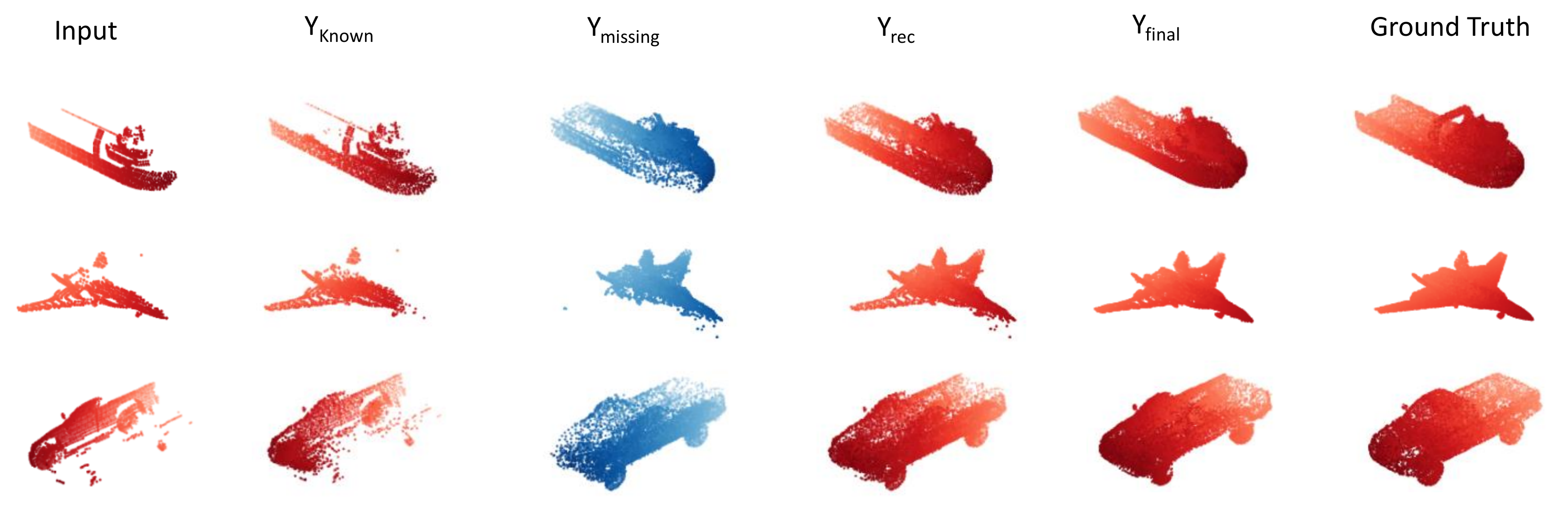}
\caption{Symmetrical characteristic during the completion.}
\label{fig:symmetry}
\end{figure}

\vspace{-0.3cm}
\subsection{Symmetrical characteristic during completion}
During the completion process, we find our network try to learn the symmetrical characteristic of the object to complete the model. As shown in Figure ~\ref{fig:symmetry}, it can be seen that the $Y_{missing}$ is close to the partial input after a proper transformation. This indicates that the details can be preserved not only  in the partial input but also in the predicted symmetrical part taking advantages of the symmetrical characteristic. 

\vspace{-0.35cm}
\subsection{Ablation Studies}
\label{sec:experiments}
Design choices involved in our network include choosing the number of level in multi-level features extraction and the combination ratio when we aggregate the known and missing part features. These parameters can influence performance so we analyze the performance of our method as a function of these parameters.

\begin{figure}[t]
\begin{minipage}[t]{0.6\linewidth} 
\centering
\includegraphics[width=0.9\textwidth]{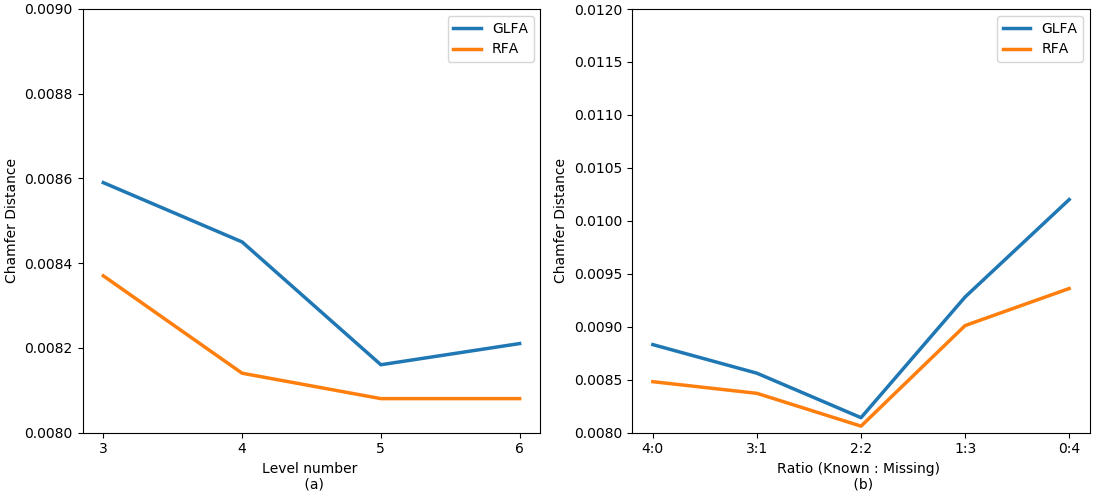}
\caption{Effects of the level number (a) and the combination ratio (b) to the evaluation metric.}
\label{fig:plot}
\end{minipage}%
\begin{minipage}[t]{.4\linewidth} 
\centering 
\includegraphics[width=1.0\textwidth]{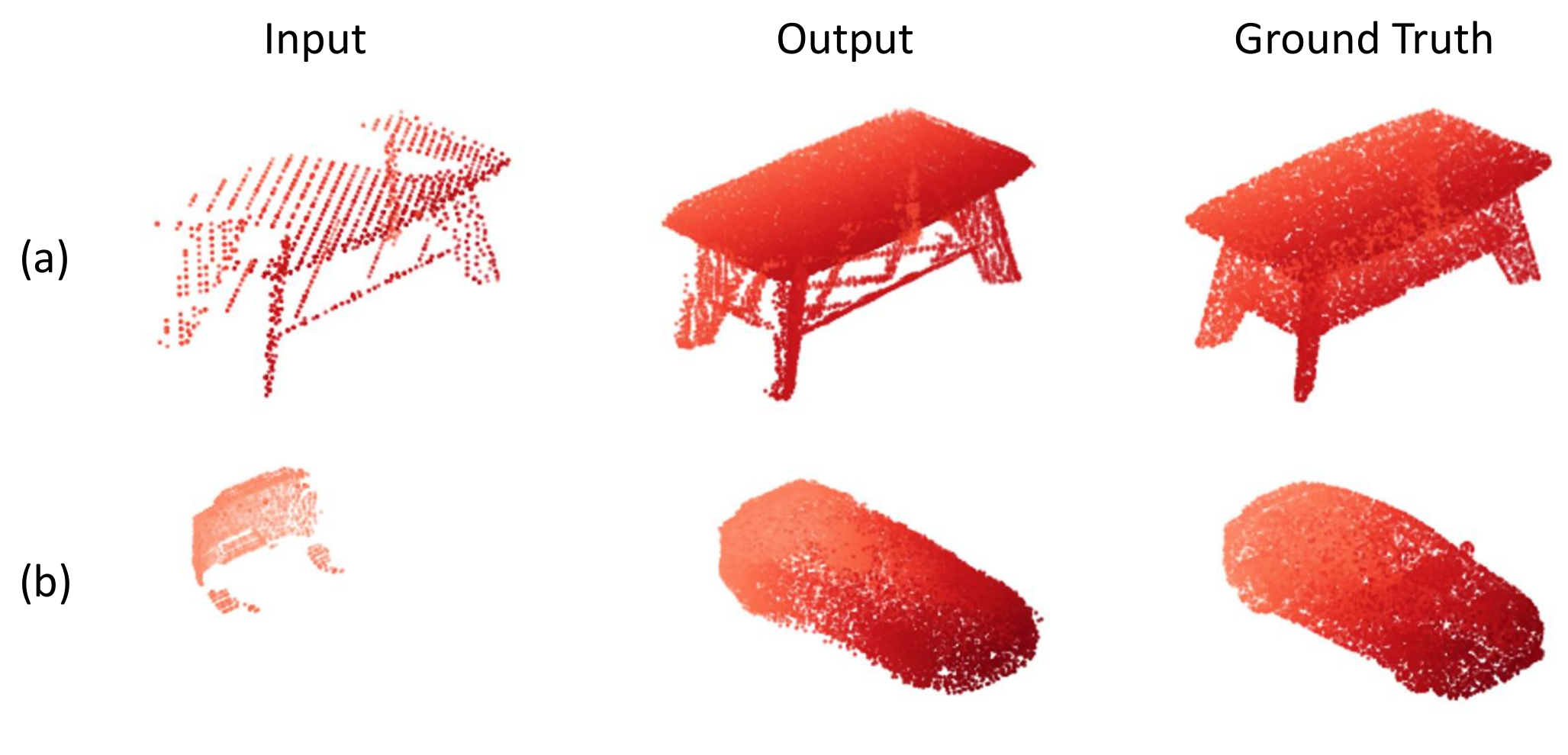}
\caption{Failure cases.}
\label{fig:failure_cases}
\end{minipage} 
\vspace{-0.6cm}
\end{figure}

\noindent{\bfseries Level number.} 
We test the performance of our network with different feature extraction level number for both NSFA-GLFA and NSFA-RFA on the testing set of known categories. Figure~\ref{fig:plot} (a) shows the Chamfer distance of our network with different level number. On the whole, the Chamfer distance reduces as the level number increases for both strategies and NSFA-RFA performs a little better than NSFA-GLFA. From level 5 to 6, the performance of NSFA-GLFA decreases. This maybe because we add much more level features to both the known and missing parts, which blurs the boundaries of the global and local features.

\noindent{\bfseries Combination ratio.}
Another important factor is the  mixture ratio. We test different mixture ratio with the baseline network. The results are shown in Figure~\ref{fig:plot} (b). When the combination ratio between known part and missing part is close to 1:1, the network achieves best performance. With ratio 1:3 and 0:4, both the performance of NSFA-RFA and NSFA-GLFA drops largely. We consider the reason is that providing few known part features makes it hard for the network to keep the details of the original model. With ratio 4:0, NSFA-RFA and NSFA-GLFA degrade to the networks directly concatenating multi-level features during the feature aggregation. 

\vspace{-0.35cm}
\subsection{Failure Cases}
\vspace{-0.2cm}
We find some failure cases during the experiments which are shown in Figure~\ref{fig:failure_cases}. They can be categorized into two cases. The first one is that, as illustrated in Figure~\ref{fig:failure_cases}(a), the partial model contains discontinuous parts (the strips under the desktop) which are caused by the view point, but the complete model has continuous shape in that area. Our network seems to regard the discontinuous parts as the details of the model and try to keep them during the completion. The other case is that the provided partial model does not have enough cues for the network to predict the details of the model as shown in Figure~\ref{fig:failure_cases}(b). Our network can recognize the model is a car but can not predict its specific details.



\vspace{-0.35cm}
\section{Conclusion}
In this work we have proposed a network for point cloud completion via with two separated feature aggregation namely GLFA and RFA, considers the existing known part and the missing part separately. RFA achieves overall better performance on known categories and GLFA shows its advantages on novel categories. Both these two strategies shows significant improvements over previous methods on both detail preservation and shape prediction.


%
%
\bibliographystyle{splncs04}
\bibliography{egbib}



\section*{Supplementary material}

## A. Appendix

In this document we provide technical details and additional discussions and results to the main paper.

In Section B, we describe the details about our network. Section C is a discussion about our work. Section D provides some additional qualitative results on real-scanned data. Section E-G show ablation studies about some choices when we design the network.

## B. Details of the network architecture

In multi-level feature extraction. we use the implementation of PointNet++. Following the notations in PointNet++, we use $(K, r, [l_1, ..., l_d])$ to represent a level with $K$ local regions of ball radius $r$, and $[l_1, ..., l_d]$ the $d$ fully connected layers with width $l_i$ $(i = 1, ..., d)$. The parameters we use are shown in Table 1.

In feature expansion and reconstruction, the separated MLPs applied for each level feature consist of fully connected layers with width [256, 128], and the shared MLP used for coordinates reconstruction is with width [64, 3]. Note that the MLP used in RFA (Residual feature aggregation) is involved in the separated MLPs which means we use the same architecture for both RFA and GLFA, but the functionality of the first fully connected layer in RFA is to transform the computed difference to the feature space.

The MLP used in the attention module is with width [16, 8, 1], which outputs a scalar as the score for each point. In local folding unit, the MLP used for generating the concatenated feature to the final coordinates is with width [512, 512, 3]. Please see our code with the link in the attached text file for more details about the implementation.

|  | Parameters | Output | Interpolated |
|---|---|---|---|
| Level 1 | $K = N, r = 0.1, mlp = [32, 32, 64]$ | $N \times 64$ | $N \times 64$ |
| Level 2 | $K = N/2, r = 0.2, mlp = [64, 64, 128]$ | $N/2 \times 128$ | $N \times 64$ |
| Level 3 | $K = N/4, r = 0.3, mlp = [128, 128, 256]$ | $N/4 \times 256$ | $N \times 64$ |
| Level 4 | $K = N/8, r = 0.4, mlp = [256, 256, 512]$ | $N/8 \times 512$ | $N \times 64$ |
| Level 5 | $K = N/16, r = 0.5, mlp = [512, 512, 1024]$ | $N/16 \times 1024$ | $N \times 64$ |
| Level 6 | $K = N/32, r = 0.6, mlp = [512, 512, 1024]$ | $N/32 \times 1024$ | $N \times 64$ |
| Global | $N = 1, mlp = [512, 512, 1024]$ | $1 \times 1024$ | $N \times 1024$ |

**Table 1.** Parameters used in multi-level feature extraction process.

## C. Is our network just doing classification?

A recent work "What Do Single-view 3D Reconstruction Networks Learn?" [1] claims that some state-of-the-art single-view 3D reconstruction methods do not

| Method | Avg | airplane | cabinet | car | chair | lamp | sofa | table | vessel |
|---|---|---|---|---|---|---|---|---|---|
| FC | 0.65 | 0.87 | 0.57 | 0.70 | 0.57 | 0.59 | 0.50 | 0.69 | 0.67 |
| Folding | 0.62 | 0.84 | 0.59 | 0.66 | 0.55 | 0.51 | 0.51 | 0.69 | 0.63 |
| TopNet | 0.67 | 0.86 | 0.61 | 0.70 | 0.59 | 0.60 | 0.56 | 0.72 | 0.71 |
| PCN | 0.69 | 0.88 | 0.65 | **0.72** | 0.62 | 0.63 | 0.58 | 0.76 | 0.69 |
| Retrieval | 0.44 | 0.71 | 0.37 | 0.38 | 0.39 | 0.42 | 0.29 | 0.49 | 0.47 |
| GLFA | 0.74 | **0.91** | 0.64 | 0.68 | 0.70 | 0.80 | 0.62 | 0.81 | 0.79 |
| RFA | **0.76** | **0.91** | **0.66** | **0.72** | **0.74** | **0.82** | **0.63** | **0.83** | **0.79** |

**Table 2.** Evaluation of all the baselines, retrieval baseline and our methods on F-score metric.

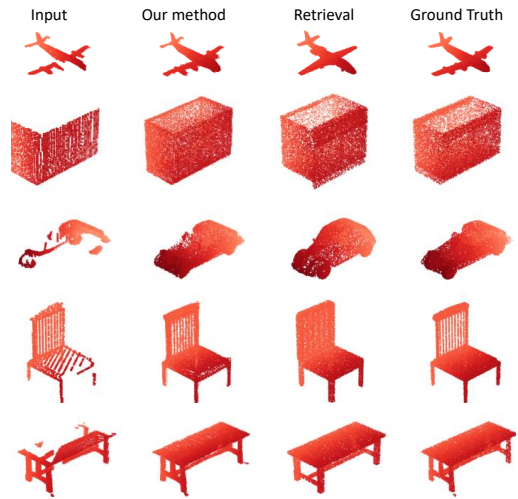

**Fig. 1.** Comparisons between our method and retrieval baseline.

actually perform reconstruction but classification. We find there are several evidences to prove our network is not just doing classification. An intuitive evidence is that our network can predict different details for every different model (see Figure 4 in our paper). If our network is just doing classification, it will predict a certain shape from a certain category. Another evidence is that instead of just using global feature to generate the complete shape, we leverage local features with proposed separated feature aggregation. It prevents the situation that the network just uses the global feature to produce a approximate shape from a certain category.

"What Do Single-view 3D Reconstruction Networks Learn?" [1] proposed the cluster and the retrieval baselines to see if the performance of the learning based single-view 3D reconstruction methods are close to just using the cluster or retrieval baseline. We intend to compare our method with these two baselines, but they are all based on single-view color image reconstruction. Following these baselines, we design another retrieval baseline based on point cloud: we retrieval

a complete ground truth point cloud from the training set which is closest to the input point cloud in terms of Chamfer Distance. We evaluate our work using F-score metric which is proposed in [1] as shown in Table 2. We can see that our method significantly outperform the retrieval baseline. Note that other baseline works also outperform the retrieval baseline. Qualitative results are also shown in Figure 1, where our method refers to NSFA-RFA. It can be observed that the retrieval method can get complete shapes without any outliers, but our method can generate more accurate results.

## D. More results on real-scanned data.

Beside real-scanned data from Kitti, we also scanned some models using the Structure Sensor to see our network performance. We show some results in Figure 2. The first column shows the color image and the second row shows the scanned partial model. We manually extract the partial object from the scanned model which is shown in the third column. The final column shows the completion results. We can see our network can recover the complete shape on these models. Note that in Figure 2(c), as our training set seems do not contains a biplane type, the produced result of the biplane might involve more noises than others.

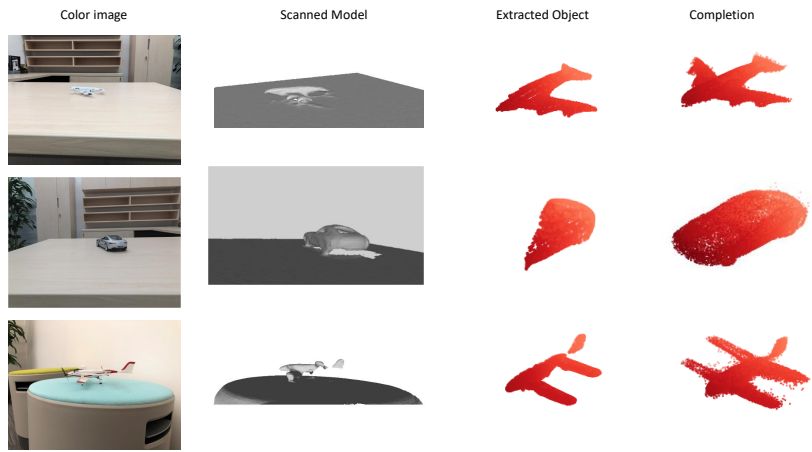

**Fig. 2.** Results on real-scanned data which is scanned by Structure Sensor.

## E. How we decide $m$ in GLFA

As mentioned in the GLFA (Global & local feature aggregation) session, we aggregate first $m$ level features to form $f_{known}$ and last $m$ level features to

express $f_{missing}$. We test the effectiveness of $m$ with networks of different feature extraction levels. The results are shown in Figure 3. Specifically, when we set $m = \lfloor \frac{n}{2} \rfloor + 1$, there are overlapped layers between $f_{known}$ and $f_{missing}$. For example, with network of 5 levels, $m$ equals 3, which means $f_{known}$ and $f_{missing}$ both involve the third-level feature. It can be seen that all the networks achieve best performance when $m = \lfloor \frac{n}{2} \rfloor + 1$. In contrast, when $m = \lfloor \frac{n}{2} \rfloor$ which indicates there are no overlapped layers between $f_{known}$ and $f_{missing}$, the performance of all networks drops dramatically. We consider the reason is that it needs at least one overlapped layer to create the correlation between $f_{known}$ and $f_{missing}$. Meanwhile, with the number of overlapped layers increasing, the performance of the networks also drops. This maybe because providing much more overlapped layers would obscure the boundary of the local feature and global feature.

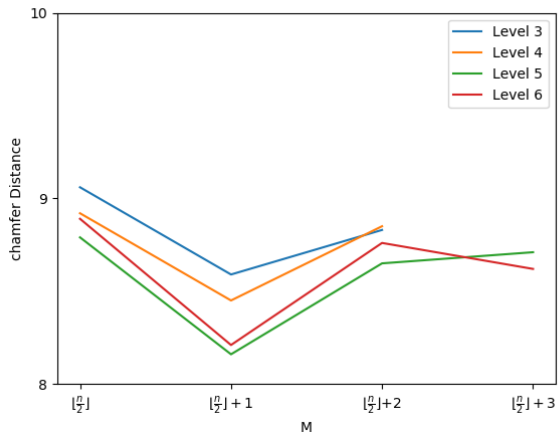

**Fig. 3.** Effect of $m$ in GLFA.

## F. The choices of $C_{missing}$

As mentioned in the Separated Feature Aggregation part, we concatenate the missing part features with the coordinates $C_{missing}$. At first, we just assign $C_{origin}$ to $C_{missing}$, but we found the network can converge faster if we set $C_{missing}$ with more proper initial coordinates. Thus we involves T-Net to represent $C_{missing}$. We compare the performance of using $C_{origin}$ and T-Net respectively, with results shown in Table . Both networks are trained in 25 epochs. We can see involves T-Net can improve the network performance on most of categories. But for NSFA-RFA on novel category, it seems that it is hard to generate proper $C_{missing}$ for residual feature aggregation. In this case, $C_{missing}$ might degrade to $C_{origin}$, where T-Net just does any transformation.

| | NSFA-RFA | | NSFA-GLFA | |
|---|---|---|---|---|
| | Known | Novel | Known | Novel |
| $C_{origin}$ | 8.56 | **10.06** | 9.48 | 11.70 |
| T-Net | **8.06** | 10.08 | **8.14** | **9.98** |

**Table 3.** Quantitative evaluation of $C_{missing}$ when using $C_{origin}$ and T-Net on Chamfer Distance multiplied by $10^4$.

## G. Evaluation of the components in loss function.

In the loss function, the repulsion term $\mathcal{L}_{rep}(Y_{coarse})$ is used for making the results to be uniformly distributed. The other $\mathcal{L}_{CD}$ components are used for intermediate and final supervision to guarantee each step results. We do evaluation by removing each term with results shown in Table 4.

| By removing | Known categories | Novel categories |
|---|---|---|
| $\mathcal{L}_{rep}(Y_{coarse})$ | 8.21 | 11.25 |
| $\mathcal{L}_{CD}(Y_{rec}, Y_{gt})$ | 8.18 | 11.05 |
| $\mathcal{L}_{CD}(Y_{coarse}, Y_{gt})$ | 8.13 | 11.40 |
| Full loss | **8.06** | **10.80** |

**Table 4.** Quantitative evaluation of loss components with NSFA-RFA on Chamfer Distance multiplied by $10^4$.


\end{document}